\documentstyle{article}
\def\report{532-96}		% fill in the report number
\title{Learning String Edit Distance% 
\thanks{Both authors are with the Department of Computer Science,
Princeton University, 35 Olden Street, Princeton, NJ 08544.  Peter Yianilos
is also with the NEC Research Institute, 4 Independence
Way, Princeton, NJ 08540.  Eric Ristad is partially supported by
Young Investigator Award IRI-9258517 from the National Science
Foundation. Email: {\tt \{ristad,pny\}@cs.princeton.edu}.}}
\author{Eric Sven Ristad \and Peter N. Yianilos}
\date{October 1996; Revised October 1997}
\newtheorem{theorem}{Theorem}
\newcommand{\proof}{\noindent{\bf Proof. }}
\def\qed{\large ~$\Box$ \normalsize}
\def\tuple<#1>{\langle #1 \rangle}
\def\argmax#1#2{\mbox{\sf argmax}_{#1}\left\{#2\right\}}
\def\ith{$i^{\rm th}$}
\begin{document}
\makeatletter
\def\abstract{ \vfil
\begin{center}{\bf Abstract}\end{center}}
\begin{titlepage}
\let\footnotesize\small \let\footnoterule\relax
\leavevmode
\vskip .7in
\begin{center}
{\Large\bf \@title \par}
\vskip 1pc
\begin{tabular}[t]{c}\@author 
\end{tabular}
 
\vskip 1pc
Research Report CS-TR-\report\par
\@date

\end{center}
\noindent
\@thanks

\begin{abstract}
In many applications, it is necessary to determine the similarity of
two strings.  A widely-used notion of string similarity is the edit
distance: the minimum number of insertions, deletions, and
substitutions required to transform one string into the other.  In
this report, we provide a stochastic model for string edit distance.
Our stochastic model allows us to learn a string edit distance
function from a corpus of examples.  We illustrate the utility of our
approach by applying it to the difficult problem of learning the
pronunciation of words in conversational speech.  In this application,
we learn a string edit distance with one fourth the error rate of the
untrained Levenshtein distance.  Our approach is applicable to any
string classification problem that may be solved using a similarity
function against a database of labeled prototypes.
\end{abstract}
\vskip 1pc

\noindent {\bf Keywords:}
string edit distance,
Levenshtein distance,
stochastic transduction,
syntactic pattern recognition,
prototype dictionary,
spelling correction,
string correction,
string similarity,
string classification,
speech recognition,
pronunciation modeling,
Switchboard corpus.
\end{titlepage}

\def\f#1{\fbox{#1}}
\section{Introduction}

In many applications, it is necessary to determine the similarity of
two strings.  A widely-used notion of string similarity is the edit
distance: the minimum number of insertions, deletions, and
substitutions required to transform one string into the other
\cite{levenshtein:66}.  In this report, we provide a stochastic model
for string edit distance.  Our stochastic interpretation allows us to
automatically learn a string edit distance from a corpus of examples.
It also leads to a variant of string edit distance, that aggregates
the many different ways to transform one string into another.  We
illustrate the utility of our approach by applying it to the difficult
problem of learning the pronunciation of words in the Switchboard
corpus of conversational speech \cite{switchboard:95}.  In this
application, we learn a string edit distance that reduces the error
rate of the untrained Levenshtein distance by a factor of four.

Let us first define our notation.  Let $A$ be a finite alphabet of
distinct symbols and let $x^T \in A^T$ denote an arbitrary string of
length $T$ over the alphabet $A$.  Then $x_i^j$ denotes the substring
of $x^T$ that begins at position $i$ and ends at position $j$.  For
convenience, we abbreviate the unit length substring $x_i^i$ as $x_i$
and the length $t$ prefix of $x^T$ as $x^t$.

A string edit distance is characterized by a triple $\tuple<A,B,c>$
consisting of the finite alphabets $A$ and $B$ and the primitive cost
function $c:E \rightarrow \Re_+$ where $\Re_+$ is the set of
nonnegative reals, $E = E_s \cup E_d \cup E_i$ is the alphabet of
primitive edit operations, $E_s = A\times B$ is the set of the
substitutions, $E_d = A\times\{\epsilon\}$ is the set of the
deletions, and $E_i = \{\epsilon\}\times B$ is the set of the
insertions.  Each such triple $\tuple<A,B,c>$ induces a distance
function $d_c:A^*\times B^*\rightarrow \Re_+$ that maps a pair of
strings to a nonnegative value.  The distance $d_c(x^t,y^v)$ between
two strings $x^t\in A^t$ and $y^v\in B^v$ is defined recursively as
\begin{equation}\label{d-c}
d_c(x^t,y^v) = \min\left\{ 
	\begin{array}{l}
	c(x_t,y_v)+d_c(x^{t-1},y^{v-1}), \\
	c(x_t,\epsilon)+d_c(x^{t-1},y^v), \\
	c(\epsilon,y_v)+d_c(x^t,y^{v-1})
	\end{array}
	\right\}
\end{equation}
where $d_c(\epsilon,\epsilon) = 0$.  The edit distance may be computed
in $O(t\cdot v)$ time using dynamic programming
\cite{masek-paterson:80,wagner-fisher:74}.  Many excellent reviews of
the string edit distance literature are available
\cite{hall-dowling:80,kukich:92,peterson:80,sankoff-kruskal:83}.
Several variants of the edit distance have been proposed, including
the constrained edit distance \cite{oomman:86} and the normalized edit
distance \cite{marzal-vidal:93}.

A stochastic interpretation of string edit distance was first provided
by Bahl and Jelinek \cite{bahl-jelinek:75}, but without an algorithm
for learning the edit costs.  The need for such a learning algorithm
is widely acknowledged
\cite{hall-dowling:80,oommen-kashyap:96,sankoff-kruskal:83}.  The
principal contribution of this report is an efficient algorithm for
learning the primitive edit costs from a corpus of examples.  To the
best of our knowledge, this is the first published algorithm to
automatically learn the primitive edit costs.  We initially
implemented a two-dimensional variant of our approach in August 1993
for the problem of classifying greyscale images of handwritten digits.

The remainder of this report consists of four sections and two
appendices.  In section~\ref{distance-section}, we define our
stochastic model of string edit distance and provide an efficient
algorithm to learn the primitive edit costs from a corpus of string
pairs.  In section~\ref{classify-section}, we provide a stochastic
model for string classification problems, and provide an algorithm to
estimate the parameters of this model from a corpus of labeled
strings.  Our techniques are applicable to any string classification
problem that may be solved using a string distance function against a
database of labeled prototypes.  In section~\ref{application-section},
we apply our modeling techniques to the difficult problem of learning
the pronunciations of words in conversational speech.

In appendix~\ref{adhoc-appendix}, we present results for the
pronunciation recognition problem in the classic nearest-neighbor
paradigm.  In appendix~\ref{alternate-appendix}, we present an
alternate model of string edit distance, which is conditioned on
string lengths.

\section{String Distance}\label{distance-section}

We model string edit distance as a memoryless stochastic transduction
between the underlying strings $A^*$ and the surface strings $B^*$.
Each step of the transduction generates either a substitution pair
$\tuple<a,b>$, a deletion pair $\tuple<a,\epsilon>$, an insertion pair
$\tuple<\epsilon,b>$, or the distinguished termination symbol $\#$
according to a probability function $\delta:E\cup\{\#\}\rightarrow [0,1]$.
Being a probability function, $\delta(\cdot)$ satisfies the following
constraints:
\begin{displaymath}
\begin{array}{ll}
 a. & \forall z\in E\cup\{\#\}\ [\ 0\leq \delta(z)\leq 1\ ] \\
 b. & \sum_{z\in E\cup\{\#\}} \delta(z) = 1 
\end{array}
\end{displaymath}
Note that the null operation $\tuple<\epsilon,\epsilon>$ is not
included in the alphabet $E$ of edit operations.

A memoryless stochastic transducer $\phi =\tuple<A,B,\delta>$
naturally induces a probability function $p(\cdot|\phi)$ on the space
$E^*\#$ of all terminated edit sequences.  This probability function
is defined by the following generation algorithm.

{\sf
\begin{tabbing}
aaa \= aaa \= aaa \= \kill
{\sc generate}($\phi$) \\
 1. \> For $n = 1$ to $\infty$ \\
 2. \> \> pick $z_n$ from $E\cup\{\#\}$ according to $\delta(\cdot)$ \\
 3. \> \> if $z_n = \#$ [ return($z^n$); ]
\end{tabbing}
}

In our intended applications, we require a probability function on
string pairs rather than on edit sequences.  In order to obtain such a
probability function, we consider a string pair to be the equivalence
class representative for all edit sequences whose yield is that pair.
Thus, the probability of a string pair is the sum of the probabilities
of all edit sequences for that string pair.  Let $\nu(z^n\#)\in
A^*\times B^*$ be the {\it yield\/} of the terminated edit sequence
$z^n\#$.  Then we define $p(x^T,y^V|\phi)$ to be the probability of
the complex event $\nu^{-1}(\tuple<x^T,y^V>)$,
\begin{equation}\label{marginal-pr}
  p(x^T, y^V | \phi) \doteq \sum_{\{z^n\#: \nu(z^n\#)=\tuple<x^T,y^V>\}} p(z^n\#|\phi)
\end{equation}
where the probability $p(z^n\#|\phi)$ of a terminated edit sequence
$z^n\in E^n$ is simply the product of the probabilities $\delta(z_i)$
of the individual edit operations because the transducer is
memoryless.

\begin{theorem}
$p(\cdot, \cdot | \phi)$ is a valid probability function on $A^*\times
B^*$ if and only if $\delta(\cdot)$ is valid and $\delta(\#)>0$.
\end{theorem}
\proof 
If $\delta(\cdot)$ is a valid probability function and $\delta(\#)>0$,
then $p(\cdot|\phi)$ is a valid probability function on the set
$E^*\#$ of all finite terminated edit sequences because $E^*\#$ is a
complete prefix-free set.  Each terminated edit sequence $z^n\#$
yields exactly one string pair $\nu(z^n\#)$.  Therefore, the set
$A^*\times B^*$ partitions the set $E^*\#$ and $p(A^*\times B^*|\phi)
= 1$.

If $\delta(\#) = 0$, then $p(z^n\#|\phi) = p(z^n|\phi)\delta(\#) = 0$
for all finite terminated edit sequences and $p(A^*\times B^*|\phi) =
0$ because all string pairs in $A^*\times B^*$ are finite.  If
$\delta(\cdot)$ is not valid, then $p(z^n\#|\phi)$ is invalid and
$p(A^*\times B^*|\phi)$ must be invalid as well.
\qed

The use of a distinguished termination symbol $\#$ in a memoryless
process entails that the probability of an edit sequence decays
exponentially with its length.  More importantly, the probability
$p(n|\phi)$ that an edit sequence will contain $n$ operations must
also decrease uniformly at an exponential rate.
\begin{displaymath}\begin{array}{lcl}
p(n|\phi) 
	& \doteq & \sum_{z^n\in E^n} p(z^n\#|\phi) \\
	& = & (1-\delta(\#))^n \delta(\#)
\end{array}\end{displaymath}
In many natural processes, such as those involving communication, the
probability of an edit sequence does not decrease uniformly.  More
probability is assigned to the medium-length messages than to the very
short messages.  As formulated, the memoryless transducer is unable to
accurately model such processes.  In
appendix~\ref{alternate-appendix}, we present an alternate
parameterization of the transducer without a termination symbol.  In
the alternate parameterization, we directly model the probability
$p(T,V)$ that the underlying string contains $T$ symbols and the
surface string contains $V$ symbols.  As a result, the probability of
the length $n$ of the underlying edit sequence need not decrease
exponentially.

The remainder of this section explains how to use the memoryless
stochastic transducer as a string edit distance.  First we use the
stochastic transducer to define two string edit distances: the Viterbi
edit distance and the stochastic edit distance.  We show how to
efficiently evaluate the joint probability of a string pair according
to a given transducer $\phi$.  This computation is necessary to
calculate the stochastic edit distance between two strings.  Next, we
explain how to optimize the parameters of a memoryless transducer on a
corpus of similar string pairs.  This computation is equivalent to
learning the primitive edit costs.  Finally, we present three variants
on the memoryless transducer, which lead to three variants of the two
string edit distances.  Subsequently, section~\ref{classify-section}
explains how to solve string classification problems using a
stochastic transducer.

 \subsection{Two Distances}\label{two-distances-section}

Our interpretation of string edit distance as a stochastic
transduction naturally leads to the following two string distances.
The first distance $d^v_\phi(\cdot,\cdot)$ is defined by the most
likely transduction between the two strings, while the second distance
$d^s_\phi(\cdot,\cdot)$ is defined by aggregating all transductions
between the two strings.

The first transduction distance $d^v_\phi(x^T,y^V)$, which we call the
{\it Viterbi edit distance\/}, is the negative logarithm of the
probability of the most likely edit sequence for the string pair
$\tuple<x^T,y^V>$.
\nocite{viterbi:67,forney:73}
\begin{equation}\label{d-phi}
  d^v_\phi(x^T,y^V) \doteq 
	- \log \argmax{\{z^n: \nu(z^n) = \tuple<x^T,y^V>\}}{p(z^n|\phi)}
\end{equation}
This distance function is identical to the string edit distance
$d_c(\cdot,\cdot)$ where the edit costs are set to the negative
logarithm of the edit probabilities, that is, where $c(z) \doteq -
\log\delta(z)$ for all $z\in E$.

The second transduction distance $d^s_\phi(x^T,y^V)$, which we call
the {\it stochastic edit distance\/}, is the negative logarithm of the
probability of the string pair $\tuple<x^T,y^V>$ according to the
transducer $\phi$.
\begin{equation}\label{dprime-phi}
  d^s_\phi(x^T,y^V) \doteq - \log p(x^T,y^V|\phi)
\end{equation}
This second distance differs from the first in that it considers the
contribution of all ways to simultaneously generate the two strings.
If the most likely edit sequence for $\tuple<x^T,y^V>$ is
significantly more likely than any of the other edit sequences, then
the two transduction distances will be nearly equal.  However, if a
given string pair has many likely generation paths, then the
stochastic distance $d^s_\phi(\cdot,\cdot)$ can be significantly less
than the Viterbi distance $d^v_\phi(\cdot,\cdot)$.

Unlike the classic edit distance $d_c(\phi,\phi)$, our two
transduction distances are never zero unless they are infinite for all
other string pairs.  Recall that the Levenshtein distance assigns zero
cost to all identity edit operations.  Therefore, an infinite number
of identity edits is less costly than even a single insert, delete, or
substitute.  The only way to obtain this property in a transduction
distance is to assign zero probability (ie., infinite cost) to all
nonidentity operations, which would assign finite distance only to
pairs of identical strings.  Note that such a transducer would still
assign linearly increasing distance to pairs of identical strings,
unlike the Levenshtein distance.

 \subsection{Evaluation}

Our generative model assigns probability to terminated edit sequences
and the string pairs that they yield.  Each pair of strings may be
generated by many different edit sequences.  Therefore we must
calculate the probability of a pair of strings by summing the
probability $p(z^n\#|\phi)$ over all the terminated edit sequences
that yield the given string pairs (\ref{marginal-pr}).

Each string pair is generated by exponentially many edit sequences,
and so it would not be feasible to evaluate the probability of a
string pair by actually summing over all its edit sequences.  The
following dynamic programming algorithm, due to Bahl and Jelinek
\cite{bahl-jelinek:75}, calculates the probability $p(x^T,y^V|\phi)$
in $O(T\cdot V)$ time and space.  At the end of the computation, the
$\alpha_{t,v}$ entry contains the probability $p(x^t,y^v|\phi)$ of the
prefix pair $\tuple<x^t,y^v>$ and $\alpha_{T,V}$ is the probability of
the entire string pair.

{\sf
\begin{tabbing}
aaa \= aaa \= aaa \= \kill
{\sc forward-evaluate}($x^T$,$y^V$,$\phi$) \\
 1. \> $\alpha_{0,0} := 1$; \\
 2. \> For $t = 0 \ldots T$ \\
 3. \> \> For $v = 0 \ldots V$ \\
 4. \> \> \> if ($v > 1 \vee t > 1$) 
	[ $\alpha_{t,v} := 0$; ] \\
 5. \> \> \> if ($v > 1$)
	[ $\alpha_{t,v}$ += $\delta(\epsilon,y_v)\alpha_{t,v-1}$; ]\\
 6. \> \> \> if ($t > 1$)
	[ $\alpha_{t,v}$ += $\delta(x_t,\epsilon)\alpha_{t-1,v}$; ]\\
 7. \> \> \> if ($v > 1 \wedge t > 1$)
	[ $\alpha_{t,v}$ += $\delta(x_t,y_v)\alpha_{t-1,v-1}$; ]\\
 8. \> $\alpha_{T,V}$ *= $\delta(\#)$; \\
 9. \> return($\alpha$);
\end{tabbing}
}

The space requirements of this algorithm may be reduced to
$O(\min(T,V))$ at some expense in clarity.

 \subsection{Estimation}

Under our stochastic model of string edit distance, the problem of
learning the edit costs reduces to the problem of estimating the
parameters of a memoryless stochastic transducer.  For this task, we
employ the powerful expectation maximization (EM) framework
\cite{baum-eagon:67,baum-etal:70,dempster-etal:77}.  An EM algorithm
is an iterative algorithm that maximizes the probability of the
training data according to the model.  See \cite{redner-walker:84} for
a review.  The applicability of EM to the problem of optimizing the
parameters of a memoryless stochastic transducer was first noted by
Bahl, Jelinek, and Mercer \cite{bahl-jelinek:75,jelinek-etal:75},
although they did not publish an explicit algorithm for this purpose.

As its name suggests, an EM algorithm consists of two steps.  In the
expectation step, we accumulate the expectation of each hidden event
on the training corpus.  In our case the hidden events are the edit
operations used to generate the string pairs.  In the maximization
step, we set our parameter values to their relative expectations on
the training corpus.

The following {\sc expectation-maximization\/}() algorithm optimizes
the parameters $\phi$ of a memoryless stochastic transducer on a
corpus $C$ = $\tuple<x^{T_1},y^{V_1}>$, $\ldots$,
$\tuple<x^{T_n},y^{V_n}>$ of $n$ training pairs.  Each iteration of
the EM algorithm is guaranteed to either increase the probability of
the training corpus or not change the model parameters.  The
correctness of our algorithm is shown in related work
\cite{ristad-yianilos:pu533-96}.

{\sf
\begin{tabbing}
aaa \= aaa \= aaa \= \kill
{\sc expectation-maximization\/}($\phi$, $C$) \\
 1. \> until convergence \\
 2. \> \> forall $z$ in $E$ [ $\gamma(z) := 0$; ] \\
 3. \> \> for $i$ = $1$ to $n$ \\
 4. \> \> \> {\sc expectation-step\/}($x^{T_i},y^{V_i},\phi,\gamma$,1); \\
 5. \> \> {\sc maximization-step\/}($\phi$,$\gamma$);
\end{tabbing}
}

The $\gamma(z)$ variable accumulates the expected number of times that
the edit operation $z$ was used to generate the string pairs in $C$.
Convergence is achieved when the total probability of the training
corpus does not change on consecutive iterations.  In practice, we
typically terminate the algorithm when the increase in the total
probability of the training corpus falls below a fixed threshold.
Alternately, we might simply perform a fixed number of iterations.

Let us now consider the details of the algorithm, beginning with the
expectation step.  First we define our forward and backward variables.
The forward variable $\alpha_{t,v}$ contains the probability
$p(x^t,y^v|\phi)$ of generating the pair $\tuple<x^t,y^v>$ of string
prefixes.  These values are calculated by the {\sc
forward-evaluate\/}() algorithm given in the preceding section.

The following {\sc backward-evaluate\/}() algorithm calculates the
backward values.  The backward variable $\beta_{t,v}$ contains the
probability $p(x_{t+1}^T, y_{v+1}^V|\phi, \tuple<t,v>)$ of generating
the terminated suffix pair $\tuple<x_{t+1}^T,y_{v+1}^V>$.  Note that
$\beta_{0,0}$ is equal to $\alpha_{T,V}$.

{\sf
\begin{tabbing}
aaa \= aaa \= aaa \= \kill
{\sc backward-evaluate}($x^T$,$y^V$,$\phi$) \\
 1. \> $\beta_{T,V}$ := $\delta(\#)$; \\
 2. \> for $t = T \ldots 0$ \\
 3. \> \> for $v = V \ldots 0$ \\
 4. \> \> \> if ($v < V \vee t < T$) [ $\beta_{t,v} := 0$; ] \\
 5. \> \> \> if ($v < V$) 
	[ $\beta_{t,v}$ += $\delta(\epsilon,y_{v+1})\beta_{t,v+1}$; ]\\
 6. \> \> \> if ($t < T$)
	[ $\beta_{t,v}$ += $\delta(x_{t+1},\epsilon)\beta_{t+1,v}$; ]\\
 7. \> \> \> if ($v < V \wedge t < T$)
	[ $\beta_{t,v}$ += $\delta(x_{t+1},y_{v+1})\beta_{t+1,v+1}$; ]\\
 8. \> return($\beta$);
\end{tabbing}
}

Recall that $\gamma(z)$ accumulates the expected number of times the
edit operation $z$ was used to generate a given the string pair.
These values are calculated by the following {\sc
expectation-step\/}() algorithm, which assumes that the $\gamma$
accumulators have been properly initialized.  The $\lambda$ argument
weights the expectation accumulation; it is used below when we learn a
string classifier.  For the purposes of this section, $\lambda$ is
always unity.

{\sf
\begin{tabbing}
aaa \= aaa \= aaa \= aaa \= \kill
{\sc expectation-step}($x^T$,$y^V$,$\phi$,$\gamma$,$\lambda$) \\
 1. \> $\alpha$ := {\sc forward-evaluate\/}($x^T$,$y^V$,$\phi$); \\
 2. \> $\beta$ := {\sc backward-evaluate\/}($x^T$,$y^V$,$\phi$); \\
 3. \> if ($\alpha_{T,V} = 0$) [ return; ]\\
 4. \> $\gamma(\#)$ += $\lambda$; \\
 5. \> for $t = 0 \ldots T$ \\
 6. \> \> for $v = 0 \ldots V$ \\
 7. \> \> \> if ($t > 0$)
	[ $\gamma(x_t,\epsilon)$ +=
		$\lambda\alpha_{t-1,v}\delta(x_t,\epsilon)\beta_{t,v}
		/ \alpha_{T,V}$; ] \\
 8. \> \> \> if ($v > 0$)
	[ $\gamma(\epsilon,y_v)$ +=
		$\lambda\alpha_{t,v-1}\delta(\epsilon,y_v)\beta_{t,v}
		/ \alpha_{T,V}$; ] \\
 9. \> \> \> if ($t > 0 \wedge v > 0$)
	[ $\gamma(x_t,y_v)$ +=
		$\lambda\alpha_{t-1,v-1}\delta(x_t,y_v)\beta_{t,v}
		/ \alpha_{T,V}$; ]
\end{tabbing}
}

Recall that $\alpha_{T,V}$ and $\beta_{0,0}$ both contain
$p(x^T,y^V|\phi)$ after lines 1 and 2, respectively.  Line 7
accumulates the posterior probability that we were in state
$\tuple<t-1,v>$ and emitted a $\tuple<x_t,\epsilon>$ deletion
operation.  Similarly, line 8 accumulates the posterior probability
that we were in state $\tuple<t,v-1>$ and emitted a
$\tuple<\epsilon,y_v>$ insertion operation.  Line 9 accumulates the
posterior probability that we were in state $\tuple<t-1,v-1>$ and
emitted a $\tuple<x_t,y_v>$ substitution operation.

Given the expectations $\gamma$ of our edit operations, the following
{\sc maximization-step\/}() algorithm updates our model parameters
$\phi$.

{\sf
\begin{tabbing}
aaa \= aaa \= aaa \= \kill
{\sc maximization-step}($\phi$,$\gamma$) \\
 1. \> $N := \gamma(\#)$; \\
 2. \> forall $z$ in $E$ [ $N$ += $\gamma(z)$; ] \\
 3. \> forall $z$ in $E$ [ $\delta(z) := \gamma(z) / N$; ] \\
 4. \> $\delta(\#) := \gamma(\#) / N$;
\end{tabbing}
}

The {\sc expectation-step\/}() algorithm accumulates the expectations
of edit operations by considering all possible generation sequences.
It is possible to replace this algorithm with the {\sc
viterbi-expectation-step\/}() algorithm, which accumulates the
expectations of edit operations by only considering the single most
likely generation sequence for a given pair of strings.  The only
change to the {\sc expectation-step\/}() algorithm would be to replace
the subroutine calls in lines 1 and 2.  Although such a learning
algorithm is arguably more appropriate to the original string edit
distance formulation, it is less suitable in our stochastic model of
string edit distance and so we do not pursue it here.

  \paragraph{Convergence.}

\def\Sac{\tuple<\verb|a|,\verb|c|>}
\def\Sbc{\tuple<\verb|b|,\verb|c|>}
\def\Da{\tuple<\verb|a|,\epsilon>}
\def\Db{\tuple<\verb|b|,\epsilon>}

The {\sc expectation-maximization\/}() algorithm given above is
guaranteed to converge to a local maximum on a given corpus $C$, by a
reduction to finite growth models
\cite{ristad-yianilos:pu533-96,yianilos:phd}.  Here we demonstrate 
that there may be multiple local maxima, and that only one of these
need be a global maxima.

Consider a transducer $\phi$ with alphabets $A =
\{\verb|a|,\verb|b|\}$ and $B = \{\verb|c|\}$ being trained on a
corpus $C$ consisting of exactly one string pair
$\tuple<\verb|abb|,\verb|cc|>$.  We restrict our attention to local
maxima that are attainable without initializing any model parameter to
zero.  Then, depending on how $\phi$ is initialized, EM may converge
to one of the following three local maxima.
\begin{center}
\begin{tabular}{c|c|c|c||r}
$\Sac$ & $\Sbc$ & $\Da$ & $\Db$ & $-\log_2 p(C|\hat{\phi})$ \\ \hline
   0 & 2/3 & 1/3 &   0 & 2.75 \\
 1/3 & 1/3 & 1/3 &   0 & 3.75 \\
 2/9 & 4/9 & 1/9 & 2/9 & 3.92 
\end{tabular}
\end{center}

The global optimum is at $\delta(\Da) = 1/3$ and $\delta(\Sbc) = 2/3$,
for which $p(C|\phi) = 4/27$ (2.75 bits).  This maxima corresponds to
the optimal edit sequence $\Da \Sbc \Sbc$, that is, to left-insert
\verb|a| and then perform two $\Sbc$ substitutions.

A second local maxima is at $\delta(\Sac) = 1/3$, $\delta(\Sbc) =
1/3$, and $\delta(\Da) = 1/3$, for which $p(C|\phi) = 2/27$ (3.75
bits).  This maxima corresponds to the following two edit sequences
each occurring with probability $1/27$:
\begin{displaymath}\begin{array}{l}
	\Sac \Sbc \Db	\\
	\Sac \Db \Sbc 
\end{array}\end{displaymath}

A third local maxima is at $\delta(\Sac) = 2/9$, $\delta(\Sbc) = 4/9$,
$\delta(\Da) = 1/9$, and $\delta(\Db) = 2/9$ for which $p(C|\phi) =
16/243$ (3.92 bits).  This maxima corresponds to the following three
edit sequences, each occurring with probability $16/729$.
\begin{displaymath}\begin{array}{l}
	\Da \Sbc \Sbc \\
	\Sac \Sbc \Db \\
	\Sac \Db \Sbc \\
\end{array}\end{displaymath}

Our experience suggests that such local maxima are not a limitation in
practice, when the training corpus is sufficiently large.

 \subsection{Three Variants}\label{variant-section}

Here we briefly consider three variants of the memoryless stochastic
transducer.  First, we explain how to reduce the number of free
parameters in the transducer, and thereby simplify the corresponding
edit cost function.  Next, we propose a way to combine different
transduction distances using the technique of finite mixture modeling.
Finally, we suggest an even stronger class of string distances that
are based on stochastic transducers with memory.  A fourth variant --
the generalization to $k$-way transduction -- appears in related work
\cite{ristad-yianilos:pu533-96,yianilos:phd}.

  \subsubsection{Parameter Tying}\label{tying-section}

In many applications, the edit cost function is simpler than the one
that we have been considering here.  The most widely used edit
distance has only four distinct costs: the insertion cost, the
deletion cost, the identity cost, and the substitution
cost.\footnote{Bunke and Csirik \cite{bunke-csirik:95} propose an even
weaker ``parametric edit distance'' whose only free parameter is a
single substitution cost $r$.  The insertion and deletion costs are
fixed to unity while the identity cost is zero.}  Although this
simplification may result in a weaker edit distance, it has the
advantage of requiring less training data to accurately learn the edit
costs.  In the statistical modeling literature, the use of such
parameter equivalence classes is dubbed parameter tying.

It is straightforward to implement arbitrary parameter tying for
memoryless stochastic transducers.  Let $\tau(z)$ be the equivalence
class of the edit operation $z$, $\tau(z)\in 2^E$, and let
$\delta(\tau(z)) = \sum_{z^\prime\in\tau(z)} \delta(z^\prime)$ be the
total probability assigned to the equivalence class $\tau(z)$.  After
maximization, we simply set $\delta(z)$ to be uniform within the total
probability $\delta(\tau(z))$ assigned to $\tau(z)$.
\begin{displaymath}
	\delta(z) := \delta(\tau(z)) / |\tau(z)|
\end{displaymath}

  \subsubsection{Finite Mixtures}\label{mixture-section}

A $k$-component mixture transducer ${\bf\phi} =
\tuple<A,B,{\bf\mu},{\bf\delta}>$ is a linear combination of $k$
memoryless transducers defined on the same alphabets $A$ and $B$.  The
mixing parameters ${\bf\mu}$ form a probability function, where
$\mu_i$ is the probability of choosing the \ith\ memoryless
transducer.  Therefore, the total probability assigned to a pair of
strings by a mixture transducer is a weighted sum over all the
component transducers.
\begin{displaymath}
	p(x^t,y^v|{\bf\phi}) = 
		\sum_{i=1}^k p(x^t,y^v|\tuple<A,B,\delta_i>)\mu_i
\end{displaymath}
A mixture transducer combines the predictions of its component
transducers in a surprisingly effective way.  Since the cost
$-\log\mu_i$ of selecting the \ith\ component of a mixture transducer
is insignificant when compared to the total cost $-\log
p(x^t,y^v|\phi_i)$ of the string pair according to the \ith\
component, the string distance defined by a mixture transducer is
effectively the minimum over the $k$ distances defined by its $k$
component transducers.

Choosing the components of a mixture transducer is more of an art than
a science.  One effective approach is to combine simpler models with
more complex models.  We would combine transducers with varying
degrees of parameter tying, all trained on the same corpus.  The
mixing parameters could be uniform, ie., $\mu_i = 1/k$, or they could
be optimized using withheld training data (cross-estimation).

Another effective approach is to combine models trained on different
corpora.  This makes the most sense if the training corpus consists of
naturally distinct sections.  In this setting, we would train a
different transducer on each section of the corpus, and then combine
the resulting transducers into a mixture model.  The mixing parameters
could be set to the relative sizes of the corpus sections, or they
could be optimized using withheld training data.  For good measure, we
could also include a transducer that was trained on the entire corpus.

  \subsubsection{Memory}\label{memory-section}

From a statistical perspective, the memoryless transducer is quite
weak because consecutive edit operations are independent.  A more
powerful model -- the stochastic transducer with memory -- would
condition the probability $\delta(z_t| z_{t-n}^{t-1})$ of generating
an edit operation $z_t$ on a finite suffix of the edit sequence that
has already been generated.  Alternately, we might condition the
probability of an edit operation $z_t$ on (a finite suffix of) the
yield $\nu(z^{t-1}))$ of the past edit sequence.  These stochastic
transducers can be further strengthened with state-conditional
interpolation \cite{jelinek-mercer:80,ristad-thomas:acl97} or by
conditioning our edit probabilities $\delta(z_t| z_{t-n}^{t-1}, s)$ on
a hidden state $s$ drawn from a finite state space.  The details of
this approach, which is strictly more powerful than the class of
transducers considered by Bahl and Jelinek \cite{bahl-jelinek:75}, are
presented in forthcoming work.

\section{String Classification}\label{classify-section}

In the preceding section, we presented an algorithm to automatically
learn a string edit distance from a corpus of similar string pairs.
Unfortunately, this algorithm cannot be directly applied to solve
string classification problems.  In a string classification problem,
we are asked to assign strings to a finite number of classes.  To
learn a string classifier, we are presented with a corpus of labeled
strings, not pairs of similar strings.  Here we present a stochastic
solution to the string classification problem that allows us to
automatically and efficiently learn a powerful string classifier from
a corpus of labeled strings.  Our approach is the stochastic analog of
nearest-neighbor techniques.

For string classification problems, we require a conditional
probability $p(w|y^v)$ that the string $y^v$ belongs to the class $w$.
This conditional may be obtained from the joint probability $p(w,y^v)$
by a straightforward application of Bayes' rule: 
$p(w|y^v) = p(w,y^v)/p(y^v)$.  In this section, we explain how to
automatically induce a strong joint probability model
$p(w,y^v|L,\phi)$ from a corpus of labeled strings, and how to use
this model to optimally classify unseen strings.

We begin by defining our model class in
section~\ref{prototype-section}.  In section~\ref{classifier-section}
we explain how to use our stochastic model to optimally classify
unseen strings.  Section~\ref{estimation-section} explains how to
estimate the model parameters from a corpus of labeled strings.

 \subsection{Hidden Prototype Model}\label{prototype-section}

We model the joint probability $p(w,y^v)$ as the marginal of the joint
probability $p(w,x^t,y^v)$ of a class $w$, an underlying prototype
$x^t$, and an observed string $y^v$
\begin{displaymath}
	p(w,y^v) = \sum_{x^t\in A^*} p(w,x^t,y^v)
.
\end{displaymath}
The prototype strings are drawn from the alphabet $A$ while the
observed strings are drawn from the alphabet $B$.  Next, we model the
joint probability $p(w,x^t,y^v)$ as a product of conditional
probabilities,
\begin{equation}\label{lexicon-model}
	p(w,x^t,y^v|\phi,L) = 
		p(w|x^t,L)
		p(x^t,y^v|\phi)
\end{equation}
where the joint probability $p(x^t,y^v|\phi)$ of a prototype $x^t$ and
a string $y^v$ is determined by a stochastic transducer $\phi$, and
the conditional probability $p(w|x^t,L)$ of a class $w$ given a
prototype $x^t$ is determined from the probabilities $p(w,x^t|L)$ of
the labeled prototypes $\tuple<w,x^t>$ in the prototype dictionary
$L$.  This model has only $O(|L|+|A\times B|)$ free parameters:
$|L|-1$ free parameters in the lexicon model $p(w,x^t|L)$ plus
$(|A|+1)\cdot(|B|+1)-1$ free parameters in the transducer $\phi$ over
the alphabets $A$ and $B$.

We considered the alternate factorization $p(w,x^t,y^v|\phi,L) =
p(y^v|x^t,\phi) p(w,x^t|L)$ but rejected it as being inconsistent with
the main thrust of our paper, which is the automatic acquisition and
use of joint probabilities on string pairs.  We note, however, that
this alternate factorization has a more natural generative
interpretation as a giant finite mixture model with $|L|$ components
whose mixing parameters are the probabilities $p(w,x^t|L)$ of the
labeled prototypes and whose component models are the conditional
probabilities $p(y^v|x^t,\phi)$ given by the transducer $\phi$ in
conjunction with the underlying form $x^t$.  This alternate
factorization suggests a number of extensions to the model, such as
the use of class-conditional transducers $p(y^v|x^t,\phi_w)$ and
intra-class parameter tying schemes.

 \subsection{Optimal Classifier}\label{classifier-section}

The conditional probability $p(w|y^V)$, in conjunction with an
application-specific utility function $\mu:W\times W\rightarrow\Re$,
defines a classifier 
\begin{displaymath}
   \hat{u} = \argmax{u\in W}{\sum_{w\in W} \mu(u|w) p(w|y^V)}
\end{displaymath}
that maximizes the expected utility of the classification, where
$\mu(u|w)$ is the utility of returning the class $u$ when we believe
that the true class is $w$.

For each string $y^v$, the minimum error rate classifier outputs
$\hat{w}$ 
\begin{displaymath}
\begin{array}{lcl}
\hat{w} & \doteq & \argmax{w}{p(w|y^v,\phi,L)} \\
	& = & \argmax{w}{p(w,y^v|\phi,L)} \\
	& = & \argmax{w}{\sum_{x^t\in A^*} p(w,x^t,y^v|\phi,L)} \\
	& = & \argmax{w}{\sum_{x^t\in L(w)} p(w,x^t,y^v|\phi,L)} \\
\end{array}
\end{displaymath}
where $L(w)$ is the set of prototype strings for the class $w$.  This
decision rule correctly aggregates the similarity between an observed
string and all prototypes for a given class.

 \subsection{Estimation}\label{estimation-section}

Given a prototype lexicon $L:W\times 2^{A^*}\rightarrow [0,1]$ and a
corpus $C = \tuple<w_1,y^{V_1}>, \ldots, \tuple<w_n,y^{V_n}>$ of
labeled strings, we estimate the parameters of our model
(\ref{lexicon-model}) using expectation maximization for finite
mixture models \cite{dempster-etal:77}.  If the prototype dictionary is
not provided, one may be constructed from the training corpus.  Our EM
algorithm will maximize the joint probability of the corpus.

{\sf
\begin{tabbing}
aaa \= aaa \= aaa \= \kill
{\sc mixture-expectation-maximization\/}($\phi$, $L$, $C$) \\
 1. \> until convergence \\
 2. \> \> forall $z$ in $E$ [ $\gamma(z) := 0$; ] \\
 3. \> \> forall $\tuple<w,x^T>$ in $L$ [ $\gamma(w,x^T) := 0$; ] \\
 4. \> \> for $i$ = $1$ to $n$ \\
 5. \> \> \> {\sc mixture-expectation-step\/}($w_i,y^{V_i},\phi,L,\gamma$); \\
 6. \> \> {\sc mixture-maximization-step\/}($\phi$,$L$,$\gamma$);
\end{tabbing}
}

Lines 2-3 initialize the $\gamma$ expectation accumulators.  In
practice, it is advisable to add a small constant to the $\gamma$
accumulators so that no probability is optimized to zero.\footnote{In
our experiments below, we initialize $\gamma(z)$ to 0 because we have
sufficient training data for the transducer.  $\gamma(w,x^T)$ is
initialized to 0.1 because our prototype dictionary is at least as
large as our training corpus.}  Lines 4-5 take an expectation step on
every labeled string in the training corpus.  Each expectation step
increments the $\gamma$ accumulators, unless $p(w_i,y^{V_i}|\phi,L)$
is zero.  Finally, line 6 updates the model parameters in $\phi$ and
$L$ based on the accumulated expectations in $\gamma$.

The heart of the EM algorithm is the {\sc 
mixture-expectation-step\/}() procedure.

{\sf
\begin{tabbing}
aaa \= aaa \= aaa \= aaa \= \kill
{\sc mixture-expectation-step\/}($w$,$y^V$,$\phi$,$L$,$\gamma$) \\
 1. \> $Z$ := $0$ \\
 2. \> forall $x^T$ in $L(w)$ \\
 3. \> \> $\alpha(x^T)$ := $L(w,x^T) / L(x^T)$; \\
 4. \> \> $\alpha(x^T)$ *= {\sc forward-evaluate\/}($x^T$,$y^V$,$\phi$); \\
 5. \> \> $Z$ += $\alpha(x^T)$; \\
 6. \> forall $x^T$ in $L(w)$ \\
 7. \> \> $\gamma(w,x^T)$ += $\alpha(x^T) / Z$; \\
 8. \> \> {\sc expectation-step\/}($x^T$,$y^V$,$\phi$,$\gamma$,$\alpha(x^T)/Z$);
\end{tabbing}
}

Lines 1-5 accumulate the posterior probabilities $p(x^T|w,y^V,\phi,L)$
for all prototypes $x^T\in L(w)$.  $p(x^T|w,y^V,\phi,L)$ is the
probability that the labeled prototype $\tuple<w,x^T>$ generated the
observed string $y^V$ with known label $w$.
\begin{displaymath}
  p(x^T|w,y^V,\phi,L) = 
	\frac{p(w,x^T,y^V|\phi,L)}{\sum_{x^T\in L(w)} p(w,x^T,y^V|\phi,L)}
\end{displaymath}
Line 3 computes $p(w|x^T,L)$ from $p(w,x^t|L)/p(x^t|L)$ while line 4
computes $p(x^T,y^V|\phi)$.  Next, line 7 accumulates expectations for
the labeled prototypes $\tuple<w,x^T>$ in $L$.  At the end of the
first loop, $Z$ holds the marginal $p(w,y^V|\phi,L)$.  The second loop
accumulates expectations for $L$ and $\phi$.  Line 7 accumulates
expectations for the labeled prototypes in $L$, in order to reestimate
the $p(w,x^t|L)$ parameters of our lexicon.  Line 8 takes a weighted
expectation step for the transducer $\phi$ on the string pair
$\tuple<x^T,y^V>$.  The weight $\alpha(x^T)/Z$ is the posterior
probability $p(x^T|w,y^V,\phi,L)$.  As a result, this learning
algorithm only trains the transducer on similar strings.

All that remains is to provide the {\sc mixture-maximization-step\/}()
algorithm, which is straightforward.

{\sf
\begin{tabbing}
aaa \= aaa \= aaa \= \kill
{\sc mixture-maximization-step}($\phi$,$L$,$\gamma$) \\
 1. \> $N := 0$; \\
 2. \> forall $\tuple<w,x^t>$ in $L$ [ $N$ += $\gamma(w,x^t)$; ] \\
 3. \> forall $\tuple<w,x^t>$ in $L$ [ $L(w,x^t)$ := $\gamma(w,x^t) / N$; ] \\
 4. \> {\sc maximization-step\/}($\phi$,$\gamma$);
\end{tabbing}
}

Note that maximizing the joint probability $p(w,y^V|\phi,L)$ is not
the same as maximizing the conditional probability $p(w|y^V,\phi,L)$.
The algorithms presented here maximize the joint probability, although
they may be straightforwardly adapted to the later objective.
Unfortunately, neither objective is the same as minimizing the error
rate, although they are closely related in practice.

Our approach to string classification has the additional virtue of
being able to learn a new class from only a single example of that
class, without any retraining.  In the case of the pronunciation
recognition problem considered below, we can learn to recognize the
pronunciations of new words from only a single example of the new
word's pronunciation.  This possibility is suggested by the superior
performance of our techniques in experiments E3, where the lexicon is
constructed from training examples only, without any human
intervention.

In appendix~\ref{adhoc-appendix} we consider another approach to the
string classification problem based on the classic ``nearest
neighbor'' decision rule.  In this ad-hoc approach, we learn a string
edit distance using all valid pairs $\tuple<x^t,y^{V_i}>$ of
underlying forms $x^t\in L(w_i)$ and surface realizations $y^{V_i}$
for each word $w_i$ in the training corpus.  For each phonetic string
$y^{S_j}$ in the testing corpus $C^\prime$, we return the word
$\hat{v}_j$ in $D$ that minimizes the string distance $d(x^t,y^{S_j})$
among all lexical entries $\tuple<v,x^t>\in L$.  Although this
approach is technically simple, it has the unfortunate property of
training the transduction distances on both similar and dissimilar
pairs of strings.  Consequently, the performance of the transduction
distances trained using this approach are not appreciably different
from the performance of the untrained Levenshtein distance.
Experimental results obtained using this ad-hoc approach are also
included in the appendix.

\section{An Application}\label{application-section}

In this section, we apply our techniques to the problem of learning
the pronunciations of words.  A given word of a natural language may
be pronounced in many different ways, depending on such factors as the
dialect, the speaker, and the linguistic environment.  We describe one
way of modeling variation in the pronunciation of words.  Let $W$ be
the set of syntactic words in a language, let $A$ be the set of
underlying phonological segments employed by the language, and let $B$
be the set of observed phonemes.  The pronouncing lexicon
$L:W\rightarrow 2^{A^*}$ assigns a small set of underlying
phonological forms to every syntactic word in the language.  Each
underlying form in $A^*$ is then mapped to a surface form in $B^*$ by
a stochastic process.  Our goal is to recognize phonetic strings,
which will require us to map each surface form to the syntactic word
for which it is a pronunciation.

We formalize this pronunciation recognition (PR) problem as follows.
The input to Pronunciation Recognition is a six-tuple
$\tuple<W,A,B,L,C,C^\prime>$ consisting of a set $W$ of syntactic
words, an alphabet $A$ of phonological segments, an alphabet $B$ of
phonetic segments, a pronouncing lexicon $L:W\rightarrow 2^{A^*}$, a
training corpus $C = \tuple<w_1,y^{V_1}>, \ldots, \tuple<w_n,y^{V_n}>$
of labeled phonetic strings, and a testing corpus $C^\prime = y^{S_1},
\ldots, y^{S_m}$ of unlabeled phonetic strings.  Each training pair
$\tuple<w_i,y^{V_i}>$ in $C$ includes a syntactic word $w_i$, $w_i\in
W$, along with a phonetic string $y^{V_i}\in B^{V_i}$.  The output is
a set of labels $v_1, \ldots, v_m$ for the phonetic strings in the
testing corpus $C^\prime$.

The pronunciation recognition problem may be reduced to the string
classification problem: the syntactic words are the classes, the
underlying forms are the prototype strings, and the surface forms are
the surface strings in need of classification.  So let us now apply
our stochastic solution to the Switchboard corpus of conversational
speech.

 \subsection{Switchboard Corpus}

The Switchboard corpus contains over 3 million words of spontaneous
telephone speech conversations \cite{switchboard:95}.  It is
considered one of the most difficult corpora for speech recognition
(and pronunciation recognition) because of the tremendous variability
of spontaneous speech.  As of Summer 1996, speech recognition
technology has a word error rate above 45\% on the Switchboard corpus.
The same speech recognition technology achieves a word error rate of
less than 5\% on read speech.

Over 200,000 words of Switchboard have been manually assigned phonetic
transcripts at ICSI using a proprietary phonetic alphabet
\cite{greenberg-etal:96}.  The Switchboard corpus also includes a
pronouncing lexicon with 71,100 entries using a modified Pronlex
phonetic alphabet (long form) \cite{pronlex:95}.  In order to make the
pronouncing lexicon compatible with the ICSI corpus of phonetic
transcripts, we removed 148 entries from the lexicon and 73,068
samples from the ICSI corpus.\footnote{From the lexicon, we removed
148 entries whose words had unusual punctuation ([\verb|<!.|]).  From
the ICSI corpus, we removed 72,257 samples that were labeled with
silence, 688 samples with an empty phonetic transcript, 88 samples
with a fragmentary transcript due to interruptions, 27 samples with
the undocumented symbol \verb|?|, and 8 samples with the undocumented
symbol \verb|!|.  Note that the symbols \verb|?| and \verb|!| are not
part of either the ICSI phonetic alphabet or the Pronlex phonetic
alphabet (long forms), and are only used in the ICSI corpus.}  After
filtering, our pronouncing lexicon had 70,952 entries for 66,284
syntactic words over an alphabet of 42 phonemes.  Our corpus had
214,310 samples -- of which 23,955 were distinct -- for 9,015
syntactic words with 43 phonemes (42 Pronlex phonemes plus a special
``silence'' symbol).

 \subsection{Four Experiments}

We conducted four sets of experiments using seven models.  In all
cases, we partitioned our corpus of 214,310 samples 9:1 into 192,879
training samples and 21,431 test samples.  In no experiment did we
adapt our probability model (\ref{lexicon-model}) to the test data.

Our seven models consist of Levenshtein distance \cite{levenshtein:66}
as well as six variants resulting from our two interpretations of
three models.\footnote{The Levenshtein distance is the minimum number
of insertions, deletions, and substitutions required to transform one
string into another.  Thus, the Levenshtein distance is a string edit
distance where the cost of all identity substitutions is zero and all
other edit costs are unity.}  Our two interpretations are the
stochastic edit distance (\ref{dprime-phi}) and the classic edit
distance (\ref{d-phi}), also called the Viterbi edit distance.  For
each interpretation, we built a tied model with only four parameters,
an untied model, and a mixture model consisting of a uniform mixture
of the tied and untied models.

The transducer parameters are initialized uniformly before training,
as are the parameters of the word model $p(w|L)$ and the conditional
lexicon model $p(x^t|w,L)$ for all entries $\tuple<w,x^t>$ in $L$.
Note that a uniform $p(w|L)$ and a uniform $p(x^t|w,L)$ are not
equivalent to a uniform $p(w,x^t|L)$ because more frequent words tend
to have more pronunciations in the lexicon.

Our four sets of experiments are determined by how we obtain our
pronouncing lexicon.  The first two experiments use the Switchboard
pronouncing lexicon.  Experiment E1 uses the full pronouncing lexicon
for all 66,284 words while experiment E2 uses the subset of the
pronouncing lexicon for the 9,015 words in the corpus.  The second two
experiments use a lexicon derived from the corpus.  Experiment E3 uses
the training corpus only to construct the pronouncing lexicon, while
experiment E4 uses the entire corpus -- both training and testing
portions -- to construct the pronouncing lexicon.  The test corpus has
512 samples whose words did not appear in the training corpus, which
lower bounds the error rate for experiment E3 to 2.4\%.

The principal difference among these four experiments is how much
information the training corpus provides about the test corpus.  In
order of increasing information, we have E3 $<$ E1 $<$ E2 $<$ E4.  In
experiment E3, the pronouncing lexicon is constructed from the
training corpus only and therefore E3 provides no direct information
about the test corpus.  In experiment E1, the pronouncing lexicon was
constructed from the entire 3m word Switchboard corpus, and therefore
E1 provides weak knowledge of the set of syntactic words that appear
in the test corpus.  In experiment E2, the pruned pronouncing lexicon
provides stronger knowledge of the set of syntactic words that
actually appear in the test corpus, as well as their most salient
phonetic forms.  In experiment E4, the pronouncing lexicon provides
complete knowledge of the set of syntactic words paired with their
actual phonetic forms in the test corpus.

The following table presents the essential characteristics of the
lexicons used in the four experiments.

\begin{center}
\begin{tabular}{l||r|r|r|r||r|r}
	& 	  &       &       & entries & novel & entries \\
	& entries & words & forms & /word   & forms & /sample \\ \hline
E1 & 70,952 & 66,284 & 64,937 & 1.070 & 2908 &  1.895 \\
E2 &  9,621 &  9,015 &  9,343 & 1.067 & 3261 &  1.267 \\
E3 & 22,140 &  8,570 & 17,880 & 2.583 & 1773 &  9.434 \\
E4 & 23,955 &  9,015 & 19,355 & 2.657 &    0 & 10.027
\end{tabular}
\end{center}

The first four fields of the table pertain to the lexicon alone.
`Entries' is the number of entries in the lexicon, `words' is the
number of unique words in the lexicon, `forms' is the number of unique
phonetic forms in the lexicon, and `entries/word' is the mean number
of entries per word.  The final two fields characterize the relation
between the lexicon and the test corpus.  `novel samples' is the
number of samples in the test corpus whose phonetic forms do not
appear in the lexicon, and `entries/sample' is the mean number of
lexical entries that exactly match the phonetic form of a sample in
the test corpus.

For each experiment, we report the fraction of misclassified samples
in the testing corpus (ie., the word error rate).  Note that the
pronouncing lexicons have many homophones.  Our decision rule
$d:B^*\rightarrow 2^L$ maps each test sample $y^{S_i}$ to a subset
$d(y^{S_i}) \subset L$ of the lexical entries.  Accordingly, we
calculate the fraction of correctly classified samples as the sum over
all test samples of the ratio of the number of correct lexical entries
in $d(y^{S_i})$ to the total number of postulated lexical entries in
$d(y^{S_i})$.  The fraction of misclassified samples is one minus the
fraction of correctly classified samples.

 \subsection{Results}\label{results-section}

Our experimental results are summarized in the following table and
figures.  The table shows the word error rate for each model at the
tenth EM iteration.  After training, the error rates of the
transduction distances are from one half to one sixth the error rate
of the untrained Levenshtein distance.  The stochastic and Viterbi
edit distances have comparable performance.  The untied and mixed
models perform better than the tied model in experiments E1, E2, and
E3.

\begin{center}
\begin{tabular}{l||r|r|r|r|r|r|r}
   & Leven- & \multicolumn{3}{c|}{Stochastic Distance} 
	    & \multicolumn{3}{c}{Viterbi Distance} \\
   & shtein & Tied & Untied & Mixed & Tied & Untied & Mixed \\ \hline
E1 & 48.04 &   20.87  &    18.61  &    18.74  &   20.87  & \f{18.58} &    18.73 \\
E2 & 33.00 &   19.56  & \f{17.14} &    17.35  &   19.63  &    17.16  &    17.35 \\
E3 & 61.87 &   14.60  &    14.29  & \f{14.28} &   14.58  &    14.29  & \f{14.28} \\
E4 & 56.35 & \f{9.34} &     9.36  &     9.36  & \f{9.34} &     9.36  &     9.36
\end{tabular}
\end{center}

The error rate for experiment E3 is bounded below by 2.4\% because the
test corpus contains 512 out-of-vocabulary samplesk in the E3
experiment.  If we discard these samples, then the E3 error rate for
the untied model would drop from 14.29\% to 12.19\%.

A sparser lexicon entails a more complex mapping between underlying
forms and surface forms.  The E3 and E4 lexicons have 2.6 entries per
word, while the E1 and E2 lexicons have only 1.1 entries per word.
Consequently, the inferior performance of the transducer in E1 and E2
relative to E3 and E4 is best explained by the statistical weakness of
a transducer without memory.  The E1 lexicon has entries for 66,284
words while the E2 lexicon has entries only for the 9,015 words that
appear in the corpus.  As a result, a significant amount of the
$p(w,x^t|L)$ probability mass is assigned to words that do not appear
in either the training or testing data in experiment E1.  This
accounts for the relative performance of the transducer in E1 and E2.

In experiment E4, the lexicon contains an entry for every sample in
the test corpus.  Since the Levenshtein distance between a surface
form (in the test corpus) and an underlying form (in the lexicon) is
minimized when the two forms are identical, we might expect the
Levenshtein distance to achieve a perfect 0\% error rate in experiment
E4, instead of its actual 56.35\% error rate.  The poor performance of
the Levenshtein distance in experiment E4 is due to the fact that the
mapping from phonetic forms to syntactic words is many-to-many in the
E4 lexicon.  Each phonetic form in the test corpus appears in 10.027
entries in the E4 lexicon, on average.  The most ambiguous phonetic
form in the test corpus, ``ah'', appears 528 times in the test corpus
and exactly matches entries for the following 62 words in the E4
lexicon.  
{\tt
\begin{quote}
a \verb|a_| all an and are at by bye don't for gaw have her high hm
huh I I'll I'm I've \verb|I_| in it know little my no of oh old on or
other ought our out pay see so that the them then there they those
though to too uh uhhuh um up us was we've what who would yeah you
\end{quote}
}
The great ambiguity of ``ah'' is due to transcription errors,
segmentation errors, and the tremendous variability of spontaneous
conversational speech.

We believe that the superior performance of our statistical techniques
in experiment E3, when compared to experiments E1 and E2, has two
significant implications.  Firstly, it raises the possibility of
obsoleting the costly process of making a pronouncing lexicon by hand.
A pronouncing lexicon that is constructed directly from actual
pronunciations offers the possibility of better performance than one
constructed in traditional ways.  Secondly, it suggests that our
techniques may be able to accurately recognize the pronunciations of
new words from only a single example of the new word's pronunciation,
without any retraining.  We simply add the new word $w$ with its
observed pronunciation $x^t$ into the pronouncing lexicon $L$, and
assign the new entry a probability $p(w,x^t|L)$ based on its observed
frequency of occurrence.  The old entries in the lexicon have their
probabilities scaled down by $1 - p(w,x^t|L)$, and the transducer
$\phi$ remains constant.

 \subsection{Credit Assignment}

Recall that our joint probability model $p(w,x^t,y^v|\phi,L)$ is
constructed from three separate models: the conditional probability
$p(w|x^t,L)$ is given by the word model $p(w|L)$ and the lexical entry
model $p(x^t|w,L)$, while the joint probability $p(x^t,y^v|\phi)$ is
given by the transducer $\phi$.  Our training paradigm simultaneously
optimizes the parameters of all three models on the training corpus.
In order to better understand the contribution of each model to the
overall success of our joint model, we repeated our experiments while
alternately holding the word and lexical entry models fixed.  In all
experiments the word model $p(w|L)$ and the lexical entry model
$p(x^t|w,L)$ are initialized uniformly.  
Our results are presented in the following four tables.

\begin{description}
\item[Fix $p(w|L)$, Fix $p(x^t|w,L)$.]\mbox{ }\\

%\begin{center}
\begin{tabular}{l||r|r|r|r|r|r|r}
   & Leven- & \multicolumn{3}{c|}{Stochastic Distance} 
	    & \multicolumn{3}{c}{Viterbi Distance} \\
   & shtein & Tied &   Untied  & Mixed & Tied  &   Untied  & Mixed \\ \hline
E1 & 48.04 & 45.16 &    42.44  & 42.54 & 45.20 & \f{42.42} & 42.53 \\
E2 & 33.00 & 31.14 & \f{28.99} & 29.16 & 31.22 &    29.01  & 29.16 \\
E3 & 61.87 & 68.98 & \f{60.12} & 64.78 & 68.98 &    60.13  & 64.77 \\
E4 & 56.35 & 64.35 & \f{54.66} & 57.61 & 64.35 & \f{54.66} & 57.61
\end{tabular}
%\end{center}

\item[Adapt $p(w|L)$, Fix $p(x^t|w,L)$.]\mbox{ }\\

%\begin{center}
\begin{tabular}{l||r|r|r|r|r|r|r}
   & Leven- & \multicolumn{3}{c|}{Stochastic Distance} 
	    & \multicolumn{3}{c}{Viterbi Distance} \\
   & shtein & Tied & Untied & Mixed & Tied & Untied & Mixed \\ \hline
E1 & 48.04 & 20.91 &    18.61  & 18.74 & 20.88 & \f{18.58} & 18.73 \\
E2 & 33.00 & 19.56 & \f{17.14} & 17.35 & 19.63 &    17.17  & 17.36 \\
E3 & 61.87 & 40.55 & \f{35.13} & 38.39 & 40.54 &    35.14  & 38.39 \\
E4 & 56.35 & 35.18 & \f{27.57} & 27.64 & 35.18 & \f{27.57} & 27.64
\end{tabular}
%\end{center}

\item[Fix $p(w|L)$, Adapt $p(x^t|w,L)$.]\mbox{ }\\

%\begin{center}
\begin{tabular}{l||r|r|r|r|r|r|r}
   & Leven- & \multicolumn{3}{c|}{Stochastic Distance} 
	    & \multicolumn{3}{c}{Viterbi Distance} \\
   & shtein & Tied & Untied & Mixed & Tied & Untied & Mixed \\ \hline
E1 & 48.04 & 48.60 &    46.85  & \f{45.69} & 48.66 &    47.07  & 45.84 \\
E2 & 33.00 & 30.99 & \f{26.67} &    28.51  & 31.06 &    26.68  & 26.80 \\
E3 & 61.87 & 42.45 & \f{36.13} &    40.34  & 42.45 &    36.14  & 40.34 \\
E4 & 56.35 & 36.86 & \f{27.51} &    34.71  & 36.86 & \f{27.51} & 34.71
\end{tabular}
%\end{center}

\item[Adapt $p(w|L)$, Adapt $p(x^t|w,L)$.]\mbox{ }\\

%\begin{center}
\begin{tabular}{l||r|r|r|r|r|r|r}
   & Leven- & \multicolumn{3}{c|}{Stochastic Distance} 
	    & \multicolumn{3}{c}{Viterbi Distance} \\
   & shtein & Tied & Untied & Mixed & Tied & Untied & Mixed \\ \hline
E1 & 48.04 &   20.87  &    18.61  &    18.74  &   20.87  & \f{18.58} &    18.73 \\
E2 & 33.00 &   19.56  & \f{17.14} &    17.35  &   19.63  &    17.16  &    17.35 \\
E3 & 61.87 &   14.60  &    14.29  & \f{14.28} &   14.58  &    14.29  & \f{14.28} \\
E4 & 56.35 & \f{9.34} &     9.36  &     9.36  & \f{9.34} &     9.36  &     9.36
\end{tabular}
%\end{center}
\end{description}

For experiment E1, a uniform word model severely reduces recognition
performance.  We believe this is because 57,269 of the 66,284 the
words in the E1 lexicon (84.4\%) do not appear in either the training
or testing corpora.  Adapting the word model reduces the effective
size of the lexicon to the 8,570 words that appear in the training
corpora, which significantly improves performance.

For experiments E1 and E2, adapting the lexical entry model has almost
no effect, simply because the average number of entries per word is
$1.07$ in the E1 and E2 lexicons.

For experiments E3 and E4, adapting the word model alone is only
slightly more effective than adapting the lexical entry model alone.
Adapting either model alone reduces the error rate by nearly one half
when compared to keeping both models fixed.  In contrast, adapting
both models together reduces the error rate by one fifth to one sixth
when compared to keeping both models fixed.  Thus, there is a
surprising synergy to adapting both models together: the improvement
is substantially larger than one might expect from the improvement
obtained from adapting the models separately.

Current speech recognition technology typically employs a sparse
pronouncing lexicon of hand-crafted underlying forms and imposes a
uniform distribution on the underlying pronunciations given the words.
When the vocabulary is large or contains many proper nouns, then the
pronouncing lexicon may be generated by a text-to-speech system
\cite{riley-etal:95}.  Our results suggest that a significant
performance improvement is possible by employing a richer pronouncing
lexicon, constructed directly from observed pronunciations, along with
an adapted lexical entry model.  

This tentative conclusion is supported by Riley and Ljolje
\cite{riley-ljolje:96}, who show an improvement in speech recognizer
performance by employing a richer pronunciation model than is
customary.  Our approach differs from their approach in three
important ways.  Firstly, our underlying pronouncing lexicon is
constructed directly from the observed pronunciations, without any
human intervention, while their underlying lexicon is obtained from a
hand-built text-to-speech system.  Secondly, our probability model
$p(y^v|w)$ assigns nonzero probability to infinitely many surface
forms, while their ``network'' probability model assigns nonzero
probability to only finitely many surface forms.  Thirdly, our use of
the underlying form $x^t$ as a hidden variable means that our model
can represent arbitrary (nonlocal) dependencies in the surface forms,
which their probability model cannot.

\section{Conclusion}

We explain how to automatically learn a string distance directly from
a corpus containing pairs of similar strings.  We also explain how to
automatically learn a string classifier from a corpus of labeled
strings.  We demonstrate the efficacy of our techniques by correctly
recognizing over 85\% of the unseen pronunciations of syntactic words
in conversational speech.  The success of our approach argues strongly
for the use of stochastic models in pattern recognition systems.

\clearpage
\appendix
\section{An Ad-Hoc Solution}\label{adhoc-appendix}

In this appendix we report experimental results for a simple but
ad-hoc solution to the pronunciation recognition problem based on the
classic ``nearest neighbor'' decision rule.  Here we learn a string
distance using all valid pairs $\tuple<x^T,y^{V_i}>$ of underlying
forms $x^T\in L(w_i)$ and surface realizations $y^{V_i}$ for each word
$w_i$ in the training corpus.  For each phonetic string $y^{S_j}$ in
the testing corpus $C^\prime$, we return the word $\hat{v}_j$ in $D$
that minimizes the string distance $d(x^t,y^{S_j})$ among all lexical
entries $\tuple<v,x^t>\in L$.

Our results are presented in the following table. The most striking
property of these results is how poorly the trained transduction
distances perform relative to the simple Levenshtein distance,
particularly when the pronouncing lexicon is derived from the corpus
(experiments E3 and E4).

\begin{table}[h]
\begin{center}
\begin{tabular}{l||c|c|c|c|c|c|c}
   & Leven- & \multicolumn{3}{c|}{Stochastic Distance} 
	    & \multicolumn{3}{c}{Viterbi Distance} \\
   & shtein & Tied & Untied & Mixed & Tied & Untied & Mixed \\ \hline
E1 & 48.04 & 48.40 & 46.81 & 46.96 & 48.39 & \fbox{46.79} & 46.94 \\
E2 & 33.00 & 33.55 & 32.58 & 31.82 & 33.69 & \fbox{31.59} & 31.81 \\
E3 & \fbox{61.87} & 63.05 & 62.28 & 62.49 & 63.13 & 62.04 & 62.47 \\
E4 & \fbox{56.35} & \fbox{56.35} & 59.01 & 57.63 & 56.35 & 59.02 & 57.69
\end{tabular}
\end{center}
\caption[]{Word error rate for seven string distance functions in four
experiments.  This table shows the word error rate after the tenth EM
iteration.  None of the transduction distances is significantly better
than the untrained Levenshtein distance in this approach.}
\end{table}

We believe that the poor performance of our transduction distances in
these experiments is due to the crudeness of the ad-hoc training
paradigm.  The handcrafted lexicon used in experiments E1 and E2
contains only 1.07 entries per syntactic word.  In contrast, the
lexicons derived from the corpus contain more than 2.5 entries per
syntactic word.  These entries can be quite dissimilar, and so our
ad-hoc training paradigm trains our transduction distances on both
similar and dissimilar strings.  The results presented in
section~\ref{results-section} confirm this hypothesis.  And the poor
results obtained here with an ad-hoc approach justify the more
sophisticated approach to string classification pursued in the body of
the report (section~\ref{classify-section}).

\clearpage
\section{Conditioning on String Lengths}\label{alternate-appendix}

In the main body of this report, we presented a probability function
on string pairs qua equivalence classes of terminated edit sequences.
In order to create a valid probability function on edit sequences, we
allowed our transducer to generate a distinguished termination symbol
$\#$.  A central limitation of that model is that the probability
$p(n|\phi)$ of an edit sequence length $n$ must decrease exponentially
in $n$.  Unfortunately, this model is poorly suited to linguistic
domains.  The empirical distribution of pronunciation lengths in the
Switchboard corpus fails to fit the exponential model.

In this appendix, we present a parameterization of the memoryless
transducer $\theta$ without a termination symbol.  This
parameterization allows us to more naturally define a probability
function $p(\cdot,\cdot|\theta,T,V)$ over all strings of lengths $T$
and $V$.  Thus, unlike the probability function defined in the main
body of this report, summing $p(\cdot,\cdot|\theta,T,V)$ over all
pairs of strings in $A^T\times B^V$ will result in unity.  This
conditional probability may be extended to joint probability
$p(x^T,y^V|\theta)$ on string pairs by means of an arbitrary joint
probability $p(T,V)$ on string lengths.
\begin{displaymath}
   p(x^T, y^V | \theta) = p(x^T, y^V | \theta,T,V) p(T,V)
\end{displaymath}

As we shall see, the approach pursued in the body of the report has
the advantage of a simpler parameterization and simpler algorithms.  A
second difference between the two approaches is that in the former
approach, the transducer learns the relative lengths of the string
pairs in the training corpus while in the current approach it cannot.
In the current approach, all knowledge about string lengths is
represented by the probability function $p(T,V)$ and not by the
transducer $\theta$.

We briefly considered an alternate parameterization of the transducer,
\begin{displaymath}
  p(z^n\#|\theta) = p(z^n|\theta) p(n)
\end{displaymath}
with an explicit distribution $p(n)$ on edit sequence lengths, that
need not assign uniformly decreasing probabilities to $n$.  The
principal disadvantage of such an approach is that it signficantly
increases the computational complexity of computing $p(x^T,y^V|\phi)$.
We can no longer collapse all partial edit sequences that generate the
same prefix $\tuple<x^t,y^v>$ of the string pair $\tuple<x^T,y^V>$
because these edit sequences may be of different lengths.  As a result
the dynamic programming table for such a model must contain $O(T\cdot
V\cdot (T+V))$ entries.  In contrast, the approach that we pursue in
this appendix only admits $O(T\cdot V)$ distinct states.

We begin by presenting an alternate parameterization of the memoryless
transducer, the transition probability $\delta(\cdot)$ is represented
as the product of the probability of choosing the type of edit
operation (insertion, deletion, or substitution) and the conditional
probability of choosing the symbol(s) used in the edit operation.
This alternate parameterization has the virtue of providing a
probability function on any set of string pairs of a given length.
Finally, we present algorithms that generate, evaluate, and learn the
parameters for finite strings, conditioned on their lengths.

 \subsection{Parameterization}

A {\it factored memoryless transducer\/} 
$\theta = \tuple<A,B,{\bf\omega},{\bf\delta}>$ 
consists of two finite alphabets
$A$ and $B$ as well as the triple 
${\bf\omega} = \tuple<\omega_d,\omega_i,\omega_s>$ 
of transition probabilities and the triple 
${\bf\delta} = \tuple<\delta_d,\delta_i,\delta_s>$ of
observation probabilities.  $\omega_s$ is the probability of
generating a substitution operation and $\delta_s(a,b)$ is the
probability of choosing the particular symbols $a$ and $b$ to
substitute.  Similarly, $\omega_d$ is the probability of generating a
deletion operation and $\delta_d(a)$ is the probability of choosing
the symbol $a$ to delete, while $\omega_i$ is the probability of
generating a insertion operation and $\delta_i(b)$ is the probability
of choosing the symbol $b$ to insert. 

The translation from our factored parameterization
$\theta=\tuple<A,B,{\bf\omega},{\bf\delta}>$ back to our unfactored
parameterization $\phi=\tuple<A,B,\delta>$ is straightforward.
\begin{displaymath}
 \begin{array}{lcl}
 \delta(a,\epsilon) & = & \omega_d \delta_d(a) \\
 \delta(\epsilon,b) & = & \omega_i \delta_i(b) \\
 \delta(a,b) & = & \omega_s \delta_s(a,b) \\
 \end{array}
\end{displaymath}

The translation from the unfactored parameterization to the factored
parameterization is also straightforward.
\begin{displaymath}
 \begin{array}{lcl}
 \omega_d & = & \sum_{e\in E_d} \delta(e) \\
 \delta_d(a) & = & \delta(a,\epsilon) / \omega_d \\[0.2cm]

 \omega_i & = & \sum_{e\in E_i} \delta(e) \\
 \delta_i(b) & = & \delta(\epsilon,b) / \omega_i \\[0.2cm]

 \omega_s & = & \sum_{e\in E_s} \delta(e) \\
 \delta_s(a,b) & = & \delta(a,b) / \omega_s \\
 \end{array}
\end{displaymath}
As explained below, the factored parameterization is necessary in
order to properly accumulate expectations when the expectation
maximization algorithm is conditioned on the string lengths.

 \subsection{Generation}

A factored memoryless transducer $\theta =
\tuple<A,B,{\bf\omega},{\bf\delta}>$ induces a probability function
$p(\cdot,\cdot|\theta,T,V)$ on the joint space $A^T\times B^V$ of all
pairs of strings of length $T$ and $V$.  This probability function is
defined by the following algorithm, which generates a string pair
$\tuple<x^T,y^V>$ from the joint space $A^T\times B^V$ according to
$p(\cdot|\theta,T,V)$.

{\sf
\begin{tabbing}
aaa \= aaa \= aaa \= \kill
{\sc generate-strings}($T$,$V$,$\theta$) \\
 1. \> initialize $t := 1$; $v := 1$; \\
 2. \> while $t\leq T$ and $v\leq V$ \\
 3. \> \> pick $\tuple<a,b>$ from $E$ according to $\delta(\cdot)$ \\
 4. \> \> if ($a\in A$) then $x_t := a$; $t := t+1$; \\
 5. \> \> if ($b\in B$) then $y_v := b$; $v := v+1$; \\
 6. \> while $t\leq T$ \\
 7. \> \> pick $a$ from $A$ according to $\delta_d(\cdot)$ \\
 8. \> \> $x_t := a$; $t := t+1$; \\
 9. \> while $v\leq V$ \\
10. \> \> pick $b$ from $B$ according to $\delta_i(\cdot)$ \\
11. \> \> $y_v := b$; $v := v+1$; \\
12. \> return($\tuple<x^T,y^V>$);
\end{tabbing}
}

The {\sc generate-strings\/}() algorithm begins by drawing edit
operations from $E$ according to the edit probability $\delta(\cdot)$
until at least one of the partial strings $x^t$ and $y^v$ is complete
[lines 2-5].  If $y^v$ is complete but $x^t$ is incomplete, then we
complete $x^t$ using symbols drawn from $A$ according to the marginal
probability $\delta_d(\cdot) = \delta(\cdot|E_d)$ [lines 6-8].
Conversely, if $x^t$ is complete but $y^v$ is incomplete, then we
complete $y^v$ using symbols drawn from $B$ according to the marginal
$\delta_i(\cdot) = \delta(\cdot|E_i)$ [lines 9-11].

 \subsection{Evaluation}

The marginal probability $p(x^T,y^V|\theta,T,V)$ of a pair of strings
is calculated by summing the joint probability
$p(x^T,y^V,z^n|\theta,T,V)$ over all the edit sequences 
that could have generated those strings
\begin{displaymath}
 \begin{array}{lcl}
 p(x^T, y^V | \theta,T,V) 
	& = & \sum_{z^n\in E^*} p(x^T, y^V, z^n | \theta,T,V) \\
	& = & \sum_{z^n\in E^*} p(x^T, y^V | \theta,T,V,z^n) p(z^n | \theta,T,V) \\
	& = & \sum_{\{z^n: \nu(z^n)=\tuple<x^T,y^V>\}} p(z^n|\theta,T,V)
 \end{array}
\end{displaymath}
because $p(x^T,y^V|\theta,T,V,z^n)$ is nonzero if and only if
$\nu(z^n)=\tuple<x^T,y^V>$.  By the definition of conditional
probability, 
\begin{displaymath} 
 p(z^n | \theta,T,V) = \prod_i p(z_i | \theta,T,V, z^{i-1})
.
\end{displaymath}
By the definition of the memoryless {\sc generate-strings\/}()
function, the conditional probability $p(z_i | \theta,T,V, z^{i-1})$
of the edit operation $z_i$ depends only on the relationship between
the string lengths $T,V$ and the state $\tuple<t,v>$ of the incomplete
edit sequence $z^{i-1}$.
\begin{equation}\label{conditional-delta}
p(z_i | \theta,T,V,\tuple<t,v>) =
	\left\{
	\begin{array}{ll}
		\omega_s\delta_s(a,b) & \mbox{ if } t<T \wedge v<V \wedge z_i=\tuple<a,b> \\
		\omega_d\delta_d(a)   & \mbox{ if } t<T \wedge v<V 
						\wedge z_i=\tuple<a,\epsilon> \\
		\omega_i\delta_i(b)   & \mbox{ if } t<T \wedge v<V 
						\wedge z_i=\tuple<\epsilon,b> \\
		\delta_d(a)	& \mbox{ if } t<T \wedge v=V \wedge z_i=\tuple<a,\epsilon> \\
		\delta_i(b)	& \mbox{ if } t=T \wedge v<V \wedge z_i=\tuple<\epsilon,b> \\
		0		& \mbox{ otherwise } \\
	\end{array}
	\right.
\end{equation}

Note that the corresponding transduction distance functions
\begin{displaymath}
 \begin{array}{lcl} \displaystyle
  d_\theta(x^T,y^V|T,V) & \doteq &
	- \log \argmax{\{z^n: \rho(z^n) = \tuple<x^T,y^V>\}}{p(z^n|\theta,T,V)} \\[0.2cm]
  d^\prime_\theta(x^T,y^V|T,V) & \doteq & -\log p(x^T,y^V|\theta,T,V)
 \end{array}
\end{displaymath}
are now conditioned on the string lengths, and therefore are
finite-valued only for strings in $A^T\times B^V$.

The following algorithms calculate the probability
$p(x^T,y^V|\theta,T,V)$ in quadratic time and space $O(T\cdot V)$.
The space requirements of the algorithm may be straightforwardly
reduced to $O(\min(T,V))$.  The only difference between these versions
and their unconditional variants in the body of the report is that
conditioning on the string lengths requires us to use the conditional
probabilities $\delta_d(\cdot)$ and $\delta_i(\cdot)$ instead of the
edit probabilities $\delta(\cdot)$ when a given hidden edit sequence
has completely generated one of the strings.

The following algorithm calculates the forward values.  The forward
variable $\alpha_{t,v}$ contains the probability $p(x^t, y^v,
\tuple<t,v>|\theta, T, V)$ of passing through the state $\tuple<t,v>$
and generating the string prefixes $x^t$ and $y^v$.

{\sf
\begin{tabbing}
aaa \= aaa \= aaa \= \kill
{\sc forward-evaluate-strings}($x^T$,$y^V$,$\theta$) \\
 1. \> $\alpha_{0,0} := 1$; \\
 2. \> for $t = 1 \ldots T$ 
	[ $\alpha_{t,0} := \omega_d\delta_d(x_t)\alpha_{t-1,0}$; ]\\
 3. \> for $v = 1 \ldots V$ 
	[ $\alpha_{0,v} := \omega_i\delta_i(y_v)\alpha_{0,v-1}$; ]\\
 4. \> for $t = 1 \ldots T-1$ \\
 5. \> \> For $v = 1 \ldots V-1$ \\
 6. \> \> \> $\alpha_{t,v} := \omega_s\delta_s(x_t,y_v)\alpha_{t-1,v-1}
		+ \omega_d\delta_d(x_t)\alpha_{t-1,v}
		+ \omega_i\delta_i(y_v)\alpha_{t,v-1}$; \\
 7. \> for $t = 1 \ldots T-1$ \\
 8. \> \> $\alpha_{t,V} := \omega_s\delta_s(x_t,y_V)\alpha_{t-1,V-1}
		+ \delta_d(x_t)\alpha_{t-1,V}
		+ \omega_i\delta_i(y_V)\alpha_{t,V-1}$; \\
 9. \> for $v = 1 \ldots V-1$ \\
10. \> \> $\alpha_{T,v} := \omega_s\delta_s(x_T,y_v)\alpha_{T-1,v-1}
		+ \omega_d\delta_d(x_T)\alpha_{T-1,v}
		+ \delta_i(y_v)\alpha_{T,v-1}$; \\
11. \> $\alpha_{T,V} := \omega_s\delta_s(x_T,y_V)\alpha_{T-1,V-1}
		+ \delta_d(x_T)\alpha_{T-1,V}
		+ \delta_i(y_v)\alpha_{T,V-1}$; \\
12. \> return($\alpha$);
\end{tabbing}
}

The following algorithm calculates the backward values.  The backward
variable $\beta_{t,v}$ contains the probability $p(x_{t+1}^T,
y_{v+1}^V|\theta, T, V, \tuple<t,v>)$ of generating the string
suffixes $x_{t+1}^T$ and $y_{v+1}^V>$ from the state $\tuple<t,v>$.

{\sf
\begin{tabbing}
aaa \= aaa \= aaa \= \kill
{\sc backward-evaluate-strings}($x^T$,$y^V$,$\theta$) \\
1. \> $\beta_{T,V} := 1$; \\
2. \> for $t = T-1 \ldots 0$ 
	[ $\beta_{t,V} := \delta_d(x_{t+1})\beta_{t+1,V}$; ]\\
3. \> for $v = V-1 \ldots 0$ 
	[ $\beta_{T,v} := \delta_i(y_{v+1})\beta_{T,v+1}$; ]\\
4. \> for $t = T-1 \ldots 0$ \\
5. \> \> for $v = V-1 \ldots 0$ \\
6. \> \> \> $\beta_{t,v} := \omega_s\delta_s(x_{t+1},y_{v+1})\beta_{t+1,v+1}
		+ \omega_d\delta_d(x_{t+1})\beta_{t+1,v}
		+ \omega_i\delta_i(y_{v+1})\beta_{t,v+1}$; \\
7. \> return($\beta$);
\end{tabbing}
}

Observe that $\alpha_{t,v}\beta_{t,v}$ is probability $p(x^T, y^V,
\tuple<t,v>|\theta, T, V)$ of generating the string pair
$\tuple<x^T,y^V>$ by an edit sequence that passes through the state
$\tuple<t,v>$.

 \subsection{Estimation}

The principal difference between the two expectation step algorithms
is that {\sc expectation-step-strings\/}() must accumulate
expectations for the ${\bf\omega}$ and ${\bf\delta}$ parameter sets
separately, via the ${\bf\chi}$ and ${\bf\gamma}$ variables,
respectively.  Due to the definition (\ref{conditional-delta}) of
$p(z_i | \theta,T,V,t,v)$ above, we may only accumulate expectations
for the ${\bf\omega}$ transition parameters when no transitions are
forced.

{\sf
\begin{tabbing}
aaa \= aaa \= aaa \= \kill
{\sc expectation-step-strings}($x^T$,$y^V$,$\theta$,${\bf\chi}$,${\bf\gamma}$) \\
 1. \> $\alpha$ := {\sc forward-evaluate-strings\/}($x^T$,$y^V$,$\theta$); \\
 2. \> $\beta$ := {\sc backward-evaluate-strings\/}($x^T$,$y^V$,$\theta$); \\
 3. \> for $t = 1 \ldots T-1$ \\
 4. \> \> for $v = 1 \ldots V-1$ \\
 5. \> \> \> $m_s$ := $\alpha_{t-1,v-1}\omega_s\delta_s(x_t,y_v)\beta_{t,v} 
		/ \alpha_{T,V}$; \\
 6. \> \> \> $\gamma_s(x_t,y_v)$ += $m_s$; 
		$\chi_s$ += $m_s$; \\
 7. \> \> \> $m_d$ := $\alpha_{t-1,v}\omega_d\delta_d(x_t)\beta_{t,v}
		/ \alpha_{T,V}$; \\
 8. \> \> \> $\gamma_d(x_t)$ += $m_d$; 
		$\chi_d$ += $m_d$; \\
 9. \> \> \> $m_i$ := $\alpha_{t,v-1}\omega_i\delta_i(y_v)\beta_{t,v}
		/ \alpha_{T,V}$; \\
10. \> \> \> $\gamma_i(y_v)$ += $m_i$;
		$\chi_i$ += $m_i$; \\
11. \> for $t = 1 \ldots T-1$
	[ $\gamma_d(x_t)$ +=
		$\alpha_{t-1,V}\delta_d(x_t)\beta_{t,V}
		/ \alpha_{T,V}$; ] \\
12. \> for $v = 1 \ldots V-1$ 
	[ $\gamma_i(y_v)$ +=
		$\alpha_{T,v-1}\delta_i(y_v)\beta_{T,v}
		/ \alpha_{T,V}$; ]
\end{tabbing}
}

Recall that that $\alpha_{T,V}$ and $\beta_{0,0}$ both contain
$p(x^T,y^V|\theta,T,V)$.  Line 5 calculates the posterior probability
that we were in state $\tuple<t-1,v-1>$ and emitted a
$\tuple<x_t,y_v>$ substitution operation.  Line 6 accumulates
expectations for the $\omega_s$ parameter in the $\chi_s$ variable,
and for the $\delta_s(x_t,y_v)$ parameter in the $\gamma_s(x_t,y_v)$
variable.  Lines 7-8 accumulate the posteriori probability that we
were in state $\tuple<t-1,v>$ and emitted a $\tuple<x_t,\epsilon>$
deletion operation.  Similarly, lines 9-10 accumulate the posteriori
probability that we were in state $\tuple<t,v-1>$ and emitted a
$\tuple<\epsilon,y_v>$ insertion operation.  Lines 11 and 12
accumulate the corresponding posteriori probabilities for forced
deletion and insertion transitions, respectively.  Note that no
expectations are accumulated for $\omega_d$ or $\omega_i$ in lines 11
and 12 because these events do not on forced transitions.

Given the expectations of our transition parameters and observation
parameters, the following {\sc maximization-step-strings\/}()
algorithm updates our model parameters.

{\sf
\begin{tabbing}
aaa \= aaa \= aaa \= \kill
{\sc maximization-step-strings}($\theta$,$\chi$,$\gamma$) \\
 1. \> $N := \chi_d + \chi_i + \chi_s$; \\
 2. \> $\omega_d := \chi_d / N$; $\omega_i := \chi_i / N$; $\omega_s := \chi_s / N$; \\
 3. \> $N_d := 0$; forall $a$ in $A$ [ $N_d += \gamma_d(a)$; ] \\
 4. \> forall $a$ in $A$ [ $\delta_d(a) := \gamma_d(a) / N_d$; ] \\
 5. \> $N_i := 0$; forall $b$ in $B$ [ $N_i += \gamma_i(b)$; ] \\
 6. \> forall $b$ in $B$ [ $\delta_i(b) := \gamma_i(b) / N_i$; ] \\
 7. \> $N_s := 0$; forall $\tuple<a,b>$ in $A\times B$ [ $N_s += \gamma_s(a,b)$; ] \\
 8. \> forall $\tuple<a,b>$ in $A\times B$ [ $\delta_s(a,b) := \gamma_s(a,b) / N_s$; ]
\end{tabbing}
}

\end{document}